\documentclass[english]{IEEEtran}
\usepackage[T1]{fontenc}
\usepackage[latin9]{inputenc}
\usepackage{float}
\usepackage{graphicx}

\makeatletter

\providecommand{\tabularnewline}{\\}
\floatstyle{ruled}
\newfloat{algorithm}{tbp}{loa}
\providecommand{\algorithmname}{Algorithm}
\floatname{algorithm}{\protect\algorithmname}

\makeatother

\usepackage{babel}
\begin{document}

\title{Finding Better Topologies for Deep Convolutional Neural Networks
by Evolution}

\author{Honglei Zhang, Serkan Kiranyaz, Moncef Gabbouj}
\maketitle
\begin{abstract}
Due to the nonlinearity of artificial neural networks, designing topologies
for deep convolutional neural networks (CNN) is a challenging task
and often only heuristic approach, such as trial and error, can be
applied. Evolutionary algorithm can solve optimization problems where
the fitness landscape is unknown. However, evolutionary algorithm
is computing resource intensive, which makes it difficult for problems
when deep CNNs are involved. In this paper we propose an evolutionary
strategy to find better topologies for deep CNNs. Incorporating the
concept of knowledge inheritance and knowledge learning, our evolutionary
algorithm can be executed with limited computing resources. We applied
the proposed algorithm in finding effective topologies of deep CNNs
for the image classification task using CIFAR-10 dataset. After the
evolution, we analyzed the topologies that performed well for this
task. Our studies verify the techniques that have been commonly used
in human designed deep CNNs. We also discovered that some of the graph
properties greatly affect the system performance. We applied the guidelines
learned from the evolution and designed new network topologies that
outperform Residual Net with less layers on CIFAR-10, CIFAR-100 and
SVHN dataset. 
\end{abstract}

\begin{IEEEkeywords}
evolutionary algorithm, deep convolutional neural network, knowledge
inheritance
\end{IEEEkeywords}

\section{Introduction}

Deep convolutional neural networks (CNN) have been one of the most
important research topics in recent years, ever since its overwhelming
victory in ImageNet challenge 2012 \cite{russakovsky2015imagenet,lecun2015deeplearning}.
Part of the success of deep CNNs owns to the improvement of computing
facilities and the availability of large annotated datasets \cite{russakovsky2015imagenet}.
Due to the nonlinearity, analytically studying neural networks is
difficult. However, scientists have found various practical techniques
to improve the performance of deep CNNs, for example: using a deeper
network structure \cite{simonyan2014verydeep,szegedy2015goingdeeper,szegedy2016rethinking};
better optimization techniques \cite{sutskever2013onthe,tieleman2012lecture};
better initialization of trainable variables \cite{glorot2010understanding};
more effective activation functions \cite{nair2010rectified,maas2013rectifier};
regularization \cite{schmidhuber2015deeplearning}; batch normalization
\cite{ioffe2015batchnormalization}. 

The network topology is one of the most important aspects that affect
the performance of a deep CNN. Over the last couple of years, numerous
new network topologies have been proposed with the target of either
improving the accuracy or reducing the computational complexity. One
indisputable trend is that deep CNNs are getting deeper and more complicated.
AlexNet \cite{krizhevsky2012imagenet} has 8 layers. VGG Net extends
the AlexNet structure to 16 and 19 layers \cite{simonyan2014verydeep}.
GoogLeNet uses inception modules and has 22 layers \cite{szegedy2014goingdeeper}.
The inception structure is further enhanced with different variations
in Inception Net \cite{szegedy2016rethinking}. Residual Net \cite{he2015deepresidual}
and DenseNet \cite{huang2016densely} use cross-layer connections
to deal with the gradient explosion/vanishing problems in training
deep CNNs. Squeeze Net \cite{iandola2016squeezenet} has a bottleneck
structure that efficiently reduces the number of trainable variables
in the network. Developing network structures is a difficult task
due to the lack of mathematical tools and theoretical understanding
of the neural networks. New network structures are always heuristically
designed and evaluated by experiments. Finding the optimal topology
of deep CNNs is a very difficult task, if by any means possible. 

Inspired by nature, evolutionary algorithms find solutions of an optimization
problem using mechanisms such as mutation, reproduction and selection
\cite{whitley2012genetic}. They can be applied to difficult problems
when the underlying landscape of the fitness function is unknown.
Ever since the emerging of artificial neural networks (ANN), evolutionary
algorithms have been used to train a neural network or finding a better
topology of neural networks \cite{branke1995evolutionary,miikkulainen2017evolving,stanley2002evolving,shafiee2016deeplearning,stanley2009ahypercubebased,zoph2017learning,zoph2016neuralarchitecture}.
Extraordinary computational demand is the major obstacle when applying
an evolutionary algorithm to optimize the network topology, because
a large number of individuals must be evaluated. Training a deep CNN
is a very time consuming and resource-intensive task because of the
large model, low convergence speed and complicated hyperparameter
tuning. Early stopping is commonly used to reduce the training time
\cite{zoph2016neuralarchitecture,stanley2002evolving}. However, this
compromise does not solve the entire problem and the number of individuals
can be evaluated is normally low. Rather than understanding the topology
of a better deep CNN, previous research mainly aimed at finding an
optimal network for a specific task. They entangled the network topology
optimization, network training and hyperparameter tuning together.
To the best of our knowledge, no detailed study has been made to understand
the impact of the network topology to the performance of a deep CNN. 

In this paper, we present our research of finding better deep CNN
topologies for the image classification task using evolutionary algorithm.
Our algorithm tries to use as less heuristic knowledge as possible.
We employ the concept of knowledge inheritance and knowledge learning,
so that the evolution can be executed in a more efficient way. We
found that the top-performing networks during the evolution have similar
patterns and substructures of the human designed networks. Taking
deep CNNs as directed graphs, we studied graph properties of the top-performing
topologies. Based on our observations, we find principles and guidance
for designing new network topologies. 

The rest of the paper is organized as the following: previous works
about evolutionary algorithms for ANNs are reviewed in Section \ref{sec:Previous-Work};
details of our evolutionary algorithm are in Section \ref{sec:Methodology};
Section \ref{sec:Evolving-on-CIFAR-10} shows the results of our evolutionary
algorithm applied to the image classification task using CIFAR-10
dataset; Section \ref{sec:Discussion} presents the analysis of the
top-performing topologies during the evolution and shows the experimental
results of some networks designed based on our analysis; and conclusion
and the future work are presented in Section \ref{sec:Conclusion}. 

\section{Previous Work \label{sec:Previous-Work}}

Ever since the impressive performance in the ImageNet challenge 2012,
deep CNNs have gained tremendous attention among researchers \cite{russakovsky2015imagenet,lecun2015deeplearning,branke1995evolutionary}.
Soon, people found that deep and complex networks perform better than
shallow and simple networks. Designing better network topologies has
been a major task to further improve the performance. Some successful
topologies of human designed networks are: LeNet \cite{lecun1998gradientbased},
AlexNet \cite{krizhevsky2012imagenet}, VGG Net \cite{szegedy2014goingdeeper},
Inception Net \cite{szegedy2016rethinking}, GoogLeNet \cite{simonyan2014verydeep},
Residual Net \cite{he2015deepresidual}, DenseNet \cite{huang2016densely}
and SqueezeNet \cite{iandola2016squeezenet}. 

Other than the heuristic approaches, evolutionary algorithms have
also been used to find better network topologies. Evolutionary algorithms
are based on Darwinian-like evolutionary process, where the following
basic rules are applied \cite{whitley2012genetic}:
\begin{itemize}
\item A large number of individuals that represent possible solutions of
a problem are evaluated.
\item The survival of an individual is decided by the fitness of the individual.
\item The offspring generated from the survived individuals are similar
but not identical to their parents.
\end{itemize}
Evolutionary algorithms can be used to optimize a fitness function
when an analytical solution is difficult to achieve. 

Evolutionary algorithms have been studied and used in ANNs for many
decades \cite{yao1999evolving}. They have been proved to be effective
in: finding weights of the connections; finding better network topologies;
finding hyperparameters such as learning rule and batch size. Two
of the most important algorithms are NeuroEvolution of Augmenting
Topologies (NEAT) \cite{stanley2002evolving} and HyperNEAT \cite{stanley2009ahypercubebased}
developed by Stanley et al. NEAT is a genetic algorithm that is based
on genetic operators such as crossover, mutation and selection. Because
there is more than one way to encode a network topology by a genetic
representation, crossover of two different genomes that represent
a same network topology will cause the competing convention problem\textemdash critical
information get lost \cite{stanley2002evolving}. To deal with this
problem, NEAT encodes a network topology using genes with historical
markings. These markings record how the network has been constructed
from the origin. To prevent the evolution being dominated by local
optimal, NEAT promotes innovations with speciation. It defines a measurement
of the distance of two genomes by the number of their excess and disjoint
genes. Individuals are speciated by a compatibility threshold. HyperNEAT
uses Compositional Pattern Production Networks (CPPNs) to decode large-scale
neural networks and applies NEAT method to find an optimal solution
by evolution. NEAT is an advanced but very complicated evolutionary
scheme. One has to carefully select the evolutionary strategies and
the parameters to get the evolution done properly. 

Miikkulainen et al. used NEAT method to optimize the network topology
and the hyperparameters together \cite{miikkulainen2017evolving}.
The network topology is defined using the concept of modules and blueprints.
The structure of the blueprints is based on human-designed network
topologies. In \cite{zoph2016neuralarchitecture}, Zoph et al. used
a RNN network as a controller to generate structure and hyperparameters
of a deep CNN. To promote a complex topology, anchor point is introduced
such that any two layers can be connected. The authors applied reinforcement
learning method to train the controller network. In \cite{ha2016hypernetworks},
the method was extended by increasing the number of hyperparameters
and generated more complex network topologies. The generated networks
achieved the state-of-the-art accuracy on CIFAR-10 and ImageNet datasets.
These evolutionary strategies are based on heuristic knowledge gained
from previous experiments. However due to the fundamental difficulties
of theoretically analyzing a deep CNN topology, one could argue that
whether the existing heuristic knowledge is correct and sufficient.
Have we already explored enough to find better topologies of deep
CNNs? 

Our study focuses on the topology of deep CNNs. The evolutionary strategies
are designed with the following principles:
\begin{itemize}
\item Evolution can be executed with limited computing resources and sufficient
number of individuals shall be evaluated.
\item Use as less heuristic human knowledge as possible.
\item Focus on the topologies of deep CNNs. Hyperparameters that are not
related to the topologies shall not be involved during the evolution.
\end{itemize}
Next we describe the details of our evolutionary algorithm. 

\section{Methodology\label{sec:Methodology}}

Let $y=f(x;\theta)$ be the decision function that a deep CNN represents,
where $x$ is input variable, $y$ is output variable and $\theta$
represents the parameters. A metric function $d(y,\hat{y})$ defines
the distance between the output variable $y$ and the ground truth
value $\hat{y}$. The target of the learning is to find the optimal
$\theta^{*}$ that minimizes the expected loss of the output, such
that
\begin{equation}
\theta^{*}=\arg\min_{\theta}E\left(d\left(f(x;\theta),\hat{y}\right)\right).\label{eq:nn_target}
\end{equation}

For a deep CNN, the network topology determines the function $f$.
Parameter $\theta$ includes all trainable variables such as weights
and bias of the neurons. With a given $f$, backpropagation is normally
used to find $\theta^{*}$ by optimizing the loss function. Our target
is to find better topologies for deep CNNs using evolutionary algorithm. 

\subsection{Network topology\label{subsec:Network-topology}}

We adopt the canonical artificial neuron model that has been commonly
used in ANNs \cite{bishop1995neuralnetworks}. Let $x_{i}$ be the
input of a neuron and $y$ be the output of it. An artificial neuron
is modeled as 
\begin{equation}
y=a\left(\sum_{i=1}^{k}w_{i}x_{i}+b\right),\label{eq:neuron_model}
\end{equation}
where $w_{i}$ is the weight for the input $x_{i}$, $b$ is the bias,
$a\left(\cdot\right)$ is the activation function and $k$ is the
number of predecessor neurons. The summation $\sum_{i=1}^{k}w_{i}x_{i}$
is called propagation since it propagates the output of the predecessor
neurons to this neuron. For a convolutional neural network, the propagation
is done in the convolutional manner \cite{lecun1998gradientbased}.
Instead of seeing all neurons in the previous layer, each neuron sees
$k$ of them, where $k$ is the size of a kernel (filter). Note that
some deep CNNs do not adopt this commonly accepted model. For example,
the propagation is separated from the activation function in pre-activation
Residual Net \cite{he2016identity}. 

We model a deep CNN by a directed graph $G(N,E)$, where $N$ is the
set of nodes and $E$ is the set of edges. We use a source node to
represent the input and a sink node to represent the output of the
network. Except the sink node and the source node, each internal node
in graph $G$ performs summation, activation and pooling operation
to the output of its predecessor nodes. We name the internal nodes
convolutional nodes (corresponding to convolutional layer in some
literatures \cite{krizhevsky2012imagenet,lecun1998gradientbased,szegedy2014goingdeeper}).
The output of a convolutional node is called a feature map. A convolutional
node consists identical neurons\textemdash they have the same activation
function, the weights and the bias. Each convolutional node has the
following model parameters: activation function $a(\cdot)$, number
of channels $C$, and the types pooling operation $P$. To focus on
the effect of a network topology, we use only ReLU as the activation
function, which is usually used in deep CNNs \cite{dahl2013improving}.
The types of pooling operation can be ``None'' or ``Max-pooling''
(of stride 2). If ``Max-pooling'' is used in a convolutional node,
the size of the output feature map is half of the size of the input
feature map. 

Except those edges that connect a convolutional node and the sink
node, the edges in graph $G$ apply weight operation in the propagation
function to the output of the predecessor nodes in a convolutional
manner. The size of the kernel is defined by $k$. If the size of
the feature map of the predecessor node is different than the input
size of the current node, proper stride will be applied. The edges
that connect a convolutional node to the sink node always operate
in a fully connected manner. 

Note, the deep CNN graph is a directed graph with one source node
and one sink node. The graph must be acyclic to guarantee that it
represents a feed-forward neural network. Every internal node in this
graph must be in one of the path between the source node and the sink
node. 

\subsection{Mutations\label{subsec:Mutations}}

Our evolution starts from the simplest network topology that contains
a source node, a sink node and a convolutional node. Every time an
individual is reproduced, a mutation is random selected from the predefined
mutations and applied to the topology of its parent. The following
mutations may happen during the evolution:
\begin{enumerate}
\item Double the number of channels of a convolutional node
\item Add a new convolutional node to the graph. Two nodes are randomly
selected and linked to the new node. By default, the kernel size of
the new edges is 3. Proper stride is applied if the sizes of the feature
maps of two connected nodes differ. 
\item Connect two nodes in the graph by a new edge. Kernel size is random
selected from $1,3,5,7,9.$ 
\item Prune an edge. After the edge is removed, the nodes and the edges
that are not in a path between the source node and the sink node are
removed from the graph. 
\item Insert a node to an edge. If the node is inserted to the edge that
connects the sink node, max-pooling is applied to the predecessor
node of the inserted node. The number of the channel of the inserted
node is same as its predecessor node.
\end{enumerate}

\subsection{Reproduction\label{subsec:Reproduction}}

NEAT incorporates sexual reproduction where crossover happens on the
chromosomes of the parents. Crossover is important to create variations
in offspring thus enhance the exploration during the evolution. However,
in nature, sexual reproduction appears much later than the asexual
reproduction. Early organisms, in particular the single cell organisms,
reproduce asexually. The genetic traits of these organisms are simple
and a few mutations are sufficient to bring variation. Since our evolution
starts from the simplest topology, we mimic the evolution of the early
organisms and apply asexual reproduction. Because it is difficult
to merge two different graphs to reproduce a new graph, any harsh
rules of combining two graphs would dramatically limit the search
space. Asexual reproduction is able to search solutions in the whole
topology space. However, the offspring may be ``close'' to its parent
thus the evolution can be slow when the graph is complicated. 

To reproduce an offspring in our evolution, a mutation is randomly
selected from one of the 5 possible mutations defined in Section \ref{subsec:Mutations}.
If the network generated by a mutation is invalid, for example the
source node and the sink node is not connected or the graph is cyclic,
another mutation is randomly selected. The procedure repeats until
a valid graph is reproduced.

\subsection{Selection\label{subsec:Selection}}

Selection is the stage during the evolution in which individuals are
selected according to its fitness and the survived individuals reproduce
offspring. For the classification task, we evaluate the fitness of
each individual by the classification accuracy. 

We use stochastic rank-proportional selection strategy as our selection
method \cite{jong2016evolutionary}. The probability that an individual
can reproduce is proportional to its rank in the whole population.
We applied Boltzmann distribution as the model and the probability
mass function is defined as:
\begin{equation}
p(k)=\frac{(1-e^{-\lambda})e^{-\lambda k}}{1-e^{\lambda N}},\label{eq:bolzmann_distribution}
\end{equation}
 where $k$ is the rank of an individual, $N$ is the size of the
population and $\lambda$ is the shape parameter that balance the
exploration and exploitation. When $\lambda$ is large, $p(k)$ becomes
flat, thus the individuals with low fitness has more chance to be
selected and the exploration is enhanced. If $\lambda$ is small,
the system will concentrate on the best-performing individuals and
will be in favor of exploitation. 

\subsection{Knowledge inheritance and knowledge learning\label{subsec:Knowledge-inheritance-and}}

The biggest challenge of using evolutionary algorithm to deep CNNs
is the difficulty of evaluating each individual, since the deep CNN
has to be trained. Even using a powerful GPU, it normally takes hours
or even days to fully train a deep CNN. This requires extremely large
computational resource and very long evolving time, since a large
number of individuals have to be evaluated. A computing platform with
hundreds of GPUs were used in previous research \cite{zoph2016neuralarchitecture,zoph2017learning}.
To deal with this problem, we incorporate the concept of knowledge
inheritance and knowledge learning. 

During the evolution, the network topology is implicitly encoded and
evolved by mutations. We consider the factors except the network topology
that impact the performance of an individual as knowledge. The fundamental
difference between knowledge and gene traits is that an individual
is free to utilize, alter and contribute to the knowledge it gains.
We further divide knowledge into two categories: inheritable knowledge
and learnable knowledge. As the name suggests, inheritable knowledge
is inherited through the evolution and is useful to the offspring
in the same evolutionary branch. The weights, bias and the other learnable
parameters of the deep CNN are inheritable knowledge. Learnable knowledge
is the information collected from the whole population and can be
beneficial to every individual. We treat learning related hyperparameters,
such as learning rate, batch size and optimization method, as the
learnable knowledge. 

We first explain how the inheritable knowledge is applied to an individual
using an example. Suppose an individual is reproduced by adding an
edge to the graph of its parent. Let $\theta$ be the inheritable
knowledge that the offspring individual receives. $\theta$ contains
weights, bias and other learnable variables of the deep CNN. Let $f_{p}(x;\theta_{p})$
be the decision function of the parent and $f_{o}(x;\theta_{p},\theta_{o})$
be the decision function of the offspring, where $\theta_{o}$ contains
weights of the newly added edge. For a neural network, it is obvious
that $f_{p}(x;\theta_{p})=f_{o}(x;\theta_{p},\theta_{o}=0)$, since
$\theta_{o}$ appears only in the summation terms of $f_{o}$. We
assume that the gradient descent method is used to find the optimum
of Eq. \ref{eq:nn_target}. Let $\theta_{p}^{*}$ be the optimal solution
for the parent individual. For any $\theta_{p}$, we have $f_{o}(x;\theta_{p}^{*},\theta_{o}=0)\le f_{o}(x;\theta_{p},\theta_{o}=0)$.
Thus, the $\theta_{p}^{*}$ is a reasonable initialization to optimize
$f_{o}$. With this observation, after the topology of an individual
is generated by mutating its parent, we assign the values from its
parents as the initialization of the learnable variables of the offspring
network. For edges that do not appear in its parent network, the weights
are randomly initialized. This can be understood as knowledge inherited
from the parent. This concept is similarly to the fine-tune techniques
that are often used in deep learning systems \cite{yosinski2014howtransferable}. 

A big challenge of using evolutionary algorithm with deep CNNs is
that the performance of a deep CNN is greatly affected by the training
related hyperparameters, such as learning rate, batch size, dropout
rate and optimization method. In practice, given a network topology,
the hyperparameters are either tuned manually \cite{dahl2013improving}
or programmatically \cite{bergstra2012randomsearch}. Previous research
treats these training related hyperparameters as variables of the
fitness function and use evolutionary algorithm to find a better solution.
We argue that the performance of a deep CNN is determined by its decision
function, but the training related hyperparameters and training procedures
affects the speed and the difficulty of finding the optimal or a suboptimal
solution. An optimization method should be robust to the surface of
a fitness function. Similarly, a good fitness function has a surface
that a solution can be easily found by an optimization method. We
consider the training related hyperparameters as the knowledge that
can be learned from the population and can be taught to an individual.
We apply a Bayesian Approach (BA) to learn this information from the
population \cite{mockus2002bayesian,bergstra2011algorithms} with
the assumption that the parameters are mutually independent. Algorithm
\ref{alg:traing_related_parameters} shows the details of finding
the optimal value of a training related hyperparameter. Note that
all training related hyperparameters are discrete in our system .

\begin{algorithm}
\textbf{initialize} equal probability to the values of a parameter

\textbf{repeat}
\begin{itemize}
\item \textbf{update} the distribution from the population
\item \textbf{sample} the value of the parameter using current distribution
\item \textbf{evaluate} the individual
\item \textbf{if} the individual is selected 
\begin{itemize}
\item \textbf{update} the population 
\end{itemize}
\end{itemize}
\caption{Bayesian approach of finding training related hyperparameters \label{alg:traing_related_parameters}}
\end{algorithm}

Note that the concepts of inheritable knowledge and learnable knowledge
are major differences of our evolutionary strategy comparing to previous
evolutionary algorithms used on deep CNNs \cite{stanley2002evolving,stanley2009ahypercubebased,he2015deepresidual,ha2016hypernetworks,zoph2017learning,miikkulainen2017evolving}. 

\subsection{Evolution \label{subsec:Evoluation}}

As mentioned in Section \ref{subsec:Mutations}, our evolution starts
from an individual with the simplest network topology that contains
the source node, the sink node and a convolutional node. Each individual
runs independently. Because of the randomness involved in training
a deep CNN, our evolution allows individuals with identical network
topology and training related hyperparameters to be reproduced. Algorithm
\ref{alg:individual_pseudo_code} shows the pseudo-code of an individual.
Before an individual finishes, it spouses new individuals according
to the computing capacity. The whole evolution procedure is terminated
manually when the fitness stops improving. 

\begin{algorithm}
\textbf{select} an individual from the population as the parent based
on Eq.\ref{eq:bolzmann_distribution}

\textbf{apply} a mutation as specified in Section \ref{subsec:Mutations}

\textbf{determine} the training related hyperparameters as described
in Algorithm \ref{alg:traing_related_parameters}

\textbf{inherit} trainable parameters from its parent

\textbf{train} and evaluate the individual 

\textbf{if} the individual is selected as described in Section \ref{subsec:Selection}:
\begin{itemize}
\item \textbf{update} the population 
\end{itemize}
\textbf{spouse} new individuals according to the computing capacity

\caption{Pseudo-code of the individuals during the evolution \label{alg:individual_pseudo_code}}
\end{algorithm}

\section{Evolving on CIFAR-10 dataset\label{sec:Evolving-on-CIFAR-10}}

We used our evolution algorithm described in Section \ref{sec:Methodology}
to find effective CNN topologies for the image classification task.
CIFAR-10 database was used because of its popularity in evaluating
the performance of deep learning algorithms \cite{krizhevsky2009learning}.
The size of the dataset is large enough for the evaluation purpose
and the size of each image is appropriate so that computational complexity
is acceptable. Due to the limitation of the computing resources that
are available to us, we have to apply certain constraints. 

\subsection{Computational constraints\label{subsec:Computational-Constraints}}
\begin{itemize}
\item Each individual is trained with one epoch. Normally hundreds of epochs
are required to train a deep CNN. It would be impossible to evaluate
a deep CNN if trained with one epoch and random initialization. For
evolutionary algorithms, the training of each individual has to be
compromised. For example, in \cite{zoph2016neuralarchitecture}, the
authors trained each individual using 20 epochs. As described in Section
\ref{subsec:Knowledge-inheritance-and}, the knowledge inheritance
can greatly reduce the difficulty of training a deep CNN. We are able
to evaluate a network when it is trained with one epoch. It should
also be noted that with this constraint, the evolution encourages
topology that learn fast. Same behavior is observed in \cite{miikkulainen2017evolving}. 
\item The number of simultaneously running individuals is limited by the
computing capacity. However, the size of the population is not affected
this constraint. Each individual runs independently and the evaluation
result is stored in a centralized database. 
\item The maximum amount of training time for each individual is fixed.
If an individual is not fully trained within the time limit, its accuracy
will be evaluated using the unfinished training result. Similar idea
is applied in \cite{stanley2005evolving}. This constraint also encourages
a topology that learns fast. 
\item The maximum amount of memory required to evaluate each individual
is fixed. If an individual cannot be evaluated within the given memory,
it will be discarded. This rule promotes a topology that consumes
less computing resources.
\end{itemize}

\subsection{Evolutionary results}

We applied the proposed evolutionary algorithm described in Section
\ref{sec:Methodology} to find better network topologies using CIFAR-10
dataset. The size of the population is 1000 and the maximum number
of simultaneously running individuals is 200. Each individual uses
one CPU core and 16GB memory. Maximum training time for each individual
is 12 hours. We trained each individual using the training set of
CIFAR-10 and calculated the validation accuracy using the validation
set. To balance the accuracy of the training set and the validation
set, we evaluate each individual by taking the mean of the training
accuracy and the validation accuracy. During the evolution, the system
evaluated over 37k individuals. 

\begin{figure}
\def\arraystretch{0.8} 
\begin{centering}
\textsf{\includegraphics[scale=0.4]{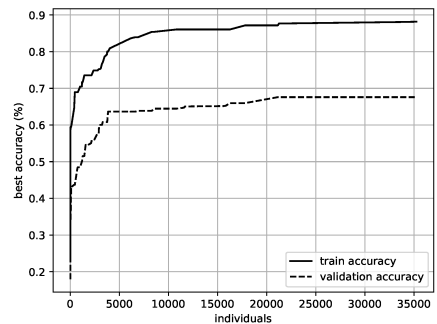}}
\par\end{centering}
\caption{Training and validation accuracy during the evolution. \label{fig:evolution_accuracy}}
\end{figure}

Fig. \ref{fig:evolution_accuracy} shows the training accuracy and
validation accuracy during the evolution. The training accuracy and
the validation accuracy get saturated as the topologies become very
complicated. Close to the end, the system took many hours to evaluate
an individual. Because of the constraint of the training time, many
individuals had been evaluated even before one epoch was finished.

\begin{figure}
\def\arraystretch{0.8} 
\begin{centering}
\textsf{}%
\begin{tabular}{cc}
\textsf{\includegraphics[scale=0.25]{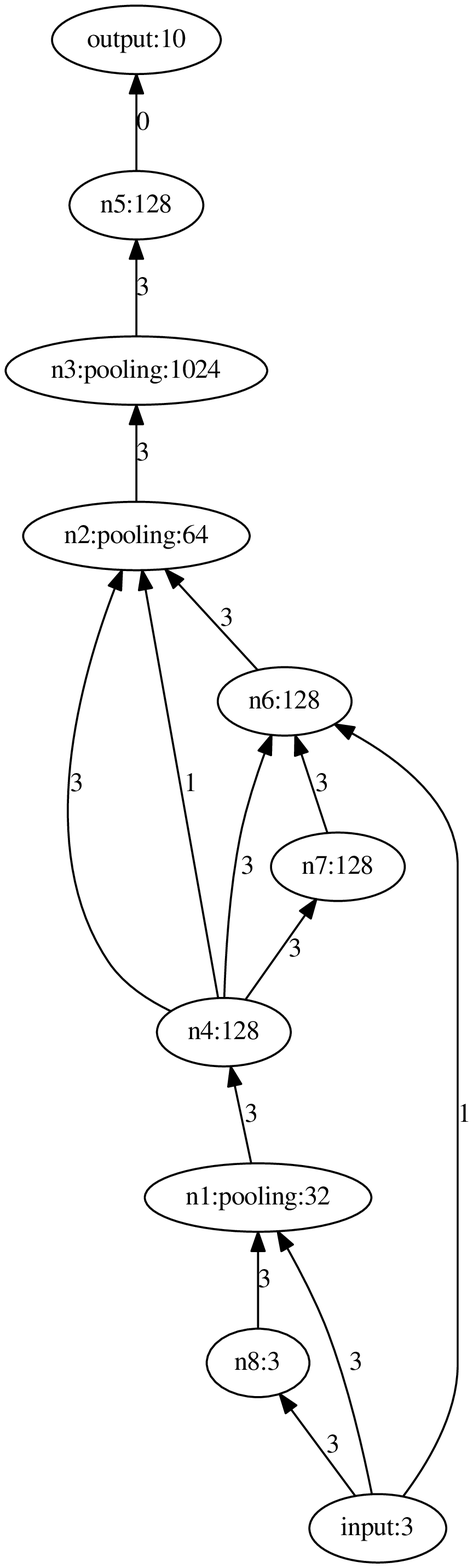}} & \includegraphics[scale=0.25]{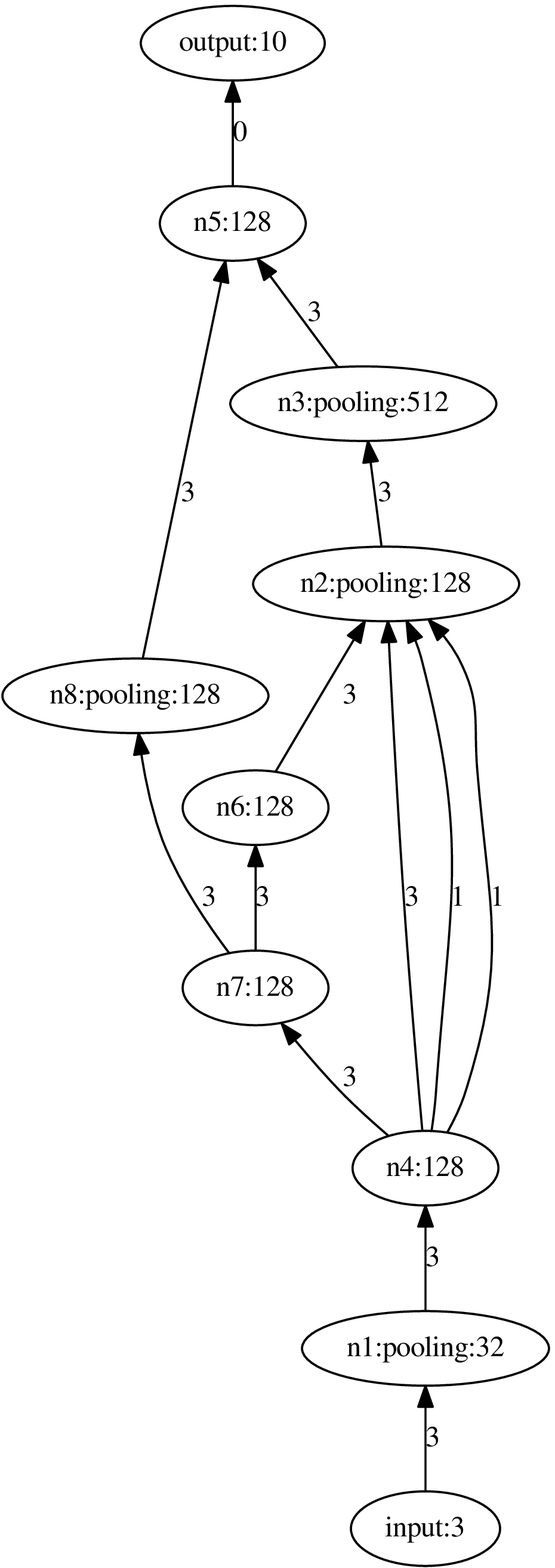}\tabularnewline
(a) & (b)\tabularnewline
 & \tabularnewline
\includegraphics[scale=0.25]{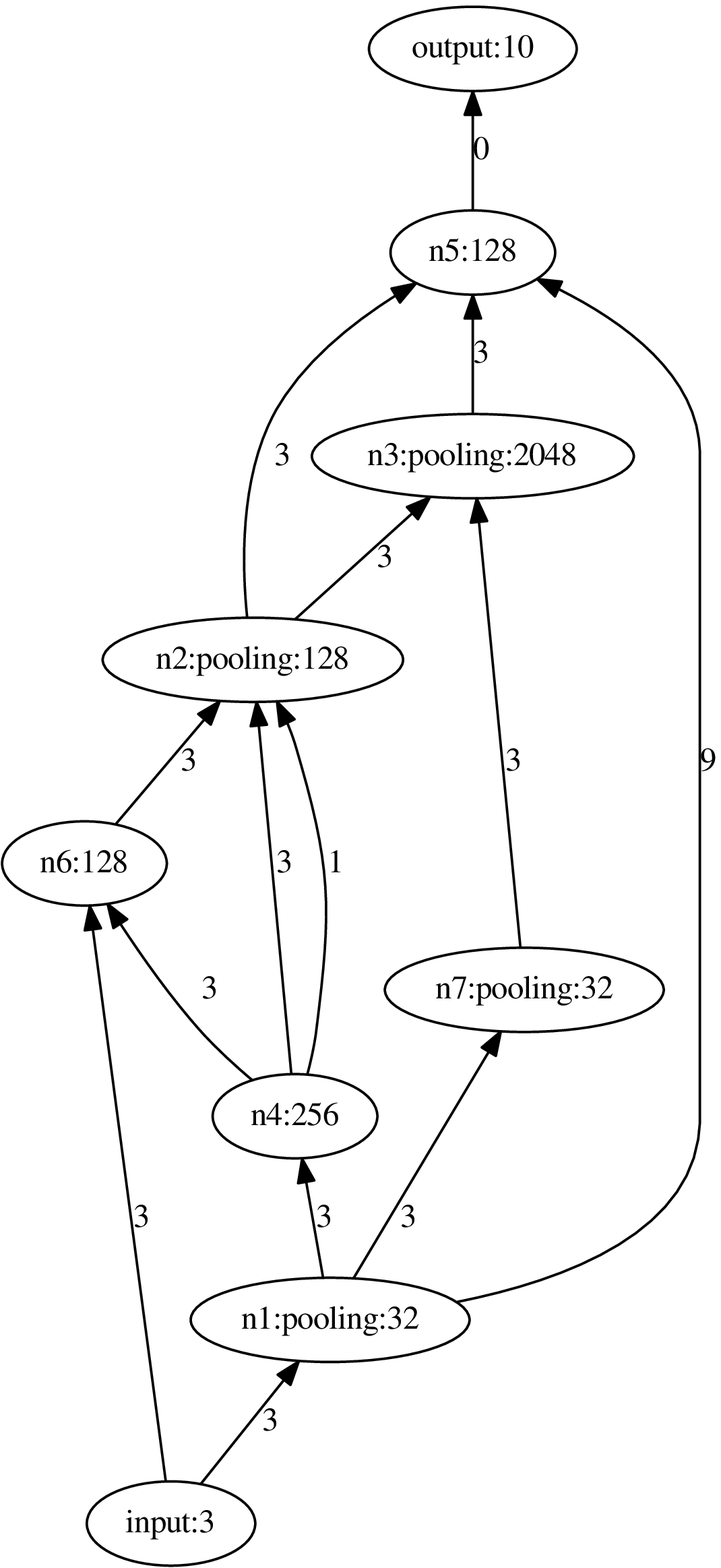} & \includegraphics[scale=0.25]{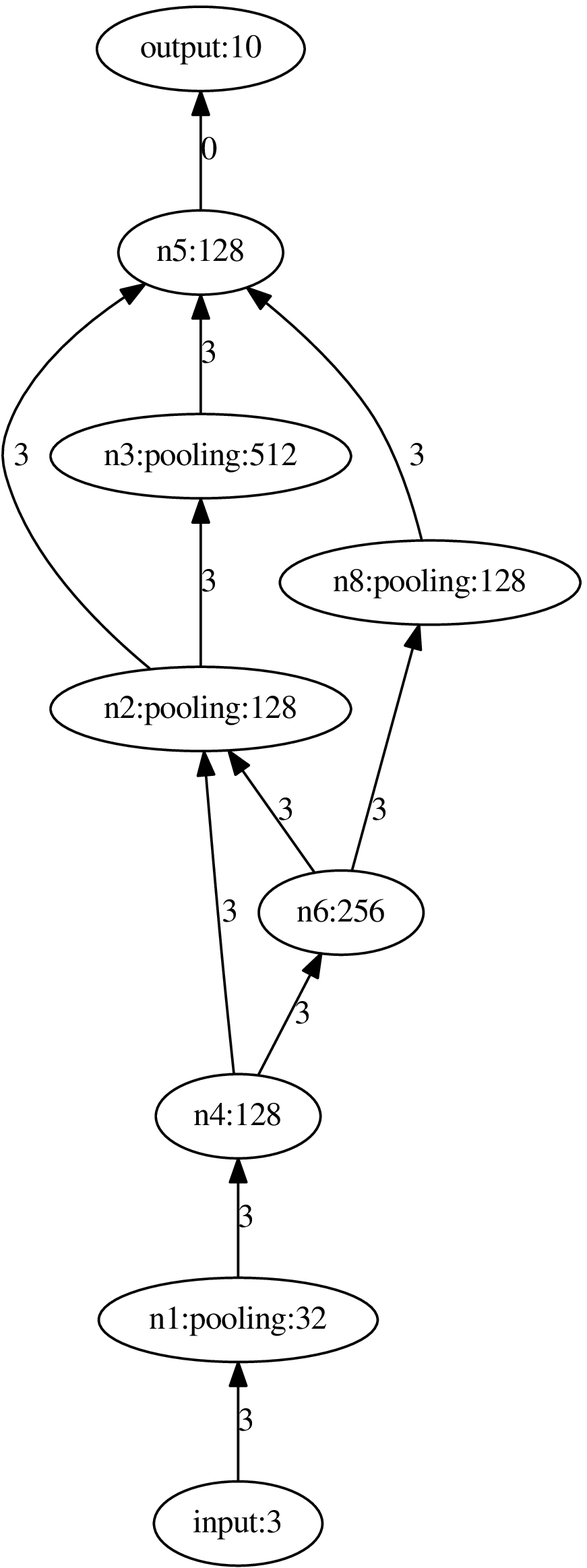}\tabularnewline
(c) & (d)\tabularnewline
\end{tabular}
\par\end{centering}
\caption{Top 4 best-performing topologies. The text in the node shows the node
name, pooling flag and the number of channels. The number next to
an edge shows the size of the kernel. \label{fig:top_networks}}
\end{figure}

Fig. \ref{fig:top_networks} shows the top 4 best-performing topologies
among all the evaluated individuals. 

\section{Discussion\label{sec:Discussion}}

\subsection{Statistics of the top-performing networks\label{subsec:Statistics-of-top}}

We first analyze some graph properties of the top-performing topologies.
For comparison, we generated the same number of networks using the
proposed evolutionary algorithm except that individuals were selected
randomly. 

We choose the best 1000 graphs from the CIFAR-10 evolved networks
and the last 1000 generated graphs from randomly evolved networks.
The average values of the following graph properties are calculated:
the number of nodes, the number of edges, graph density, graph algebraic
connectivity, average values of the number of channels (convolutional
nodes), the longest distance from the source node to the sink node
(in short, longest distance), the shortest distance from the source
node to the sink node (in short, shortest distance). Refer \cite{newman2010networks}
for details of these graph properties. Note that the longest distance
of a deep CNN is also called depth, because the early deep CNN topologies
have layered structure. Table \ref{tab:evolved_graph_properties}
shows the comparison of these graph properties between the CIFAR-10
evolved networks and the randomly evolved networks. It also shows
the average number of the generations of the individuals.

\begin{table}
\caption{Average values of some graph properties of the CIFAR-10 evolved networks
and randomly evolved networks \label{tab:evolved_graph_properties}}

\def\arraystretch{0.9} 
\centering{}%
\begin{tabular}{ccc}
\hline 
 & CIFAR-10 evolved & randomly evolved \tabularnewline
\hline 
\hline 
nodes & 9.66 & 6.515\tabularnewline
edges & 13.1 & 9.12\tabularnewline
density & 0.159 & 0.270\tabularnewline
connectivity & 0.396 & 0.721\tabularnewline
channels & 21.4 & 3.81\tabularnewline
longest dist. & 8.67 & 5.70\tabularnewline
shortest dist. & 5.60 & 3.73\tabularnewline
 generations & 33.2 & 11.8\tabularnewline
\hline 
\end{tabular}
\end{table}

\subsection{Network growth pattern}

As the statistics in Table \ref{tab:evolved_graph_properties} show,
the number of the nodes, the number of the edges and the average channels
of the CIFAR-10 evolved networks is significantly greater than those
of the randomly evolved networks. This indicates that bigger networks\textemdash more
nodes, more edges and larger number of channels\textemdash perform
better than smaller networks, thus they have better chance to be selected
during the evolution. The number of generations of the CIFAR-10 evolved
networks is also clearly greater the random evolved network. It indicates
that there is a dominant trend towards a better network topology.
The individuals with a better network topology consistently outperform
other individuals. Thus this branch gets continuously evolved and
the number of generations grows rapidly. 

It should also be noted that the density and the algebraic connectivity
of the CIFAR-10 evolved network is smaller that the randomly evolved
networks. This indicates that adding new edges does not greatly improve
the performance comparing to adding new nodes, in particular in the
early stage of the evolution. Next we try to understand the how the
network grows during the evolution and try to explain the differences
of the other properties.

\subsubsection{Going deeper is the first priority}

\begin{figure}
\def\arraystretch{0.8} 
\setlength{\tabcolsep}{2pt} 
\begin{centering}
\textsf{}%
\begin{tabular}{ccc}
\textsf{\includegraphics[scale=0.25]{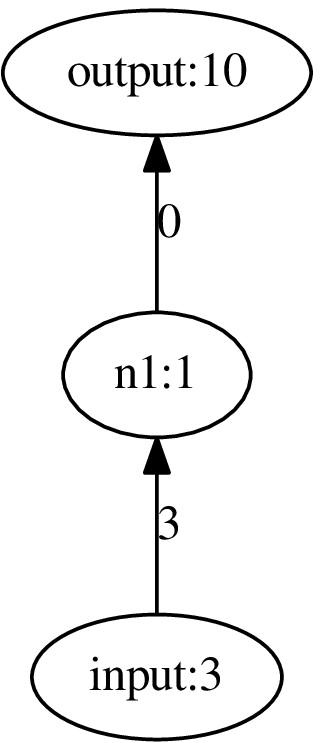}} & \includegraphics[scale=0.25]{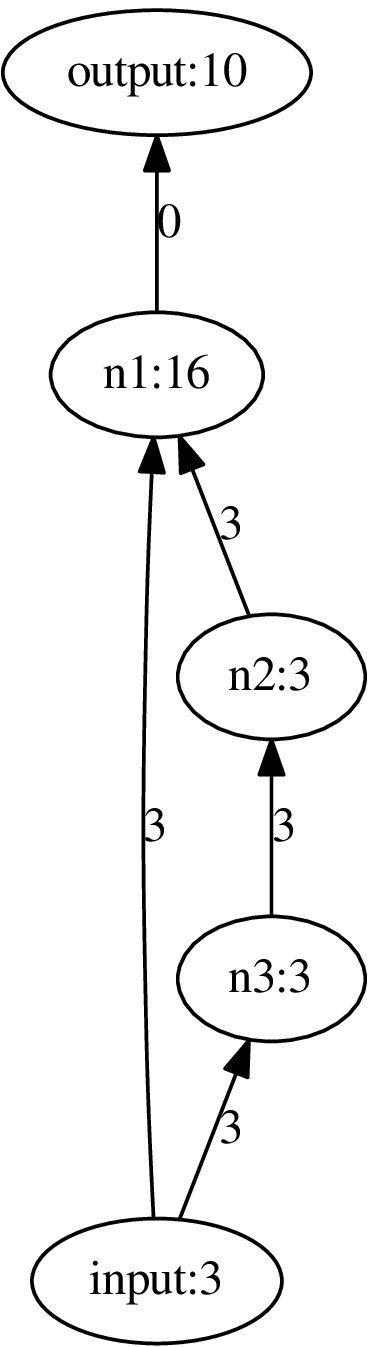} & \includegraphics[scale=0.25]{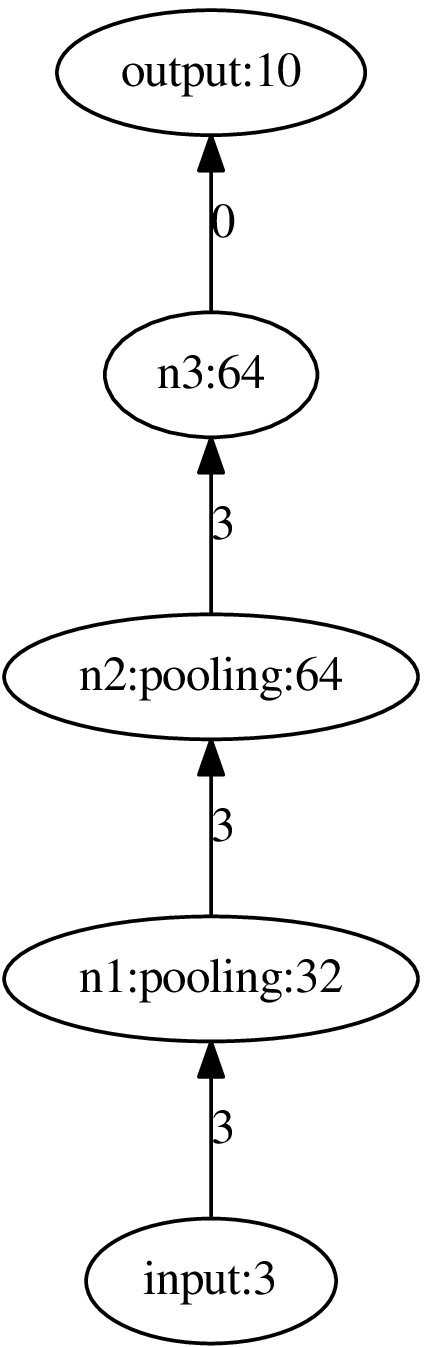}\tabularnewline
(1) & (7) & (13)\tabularnewline
 &  & \tabularnewline
\includegraphics[scale=0.25]{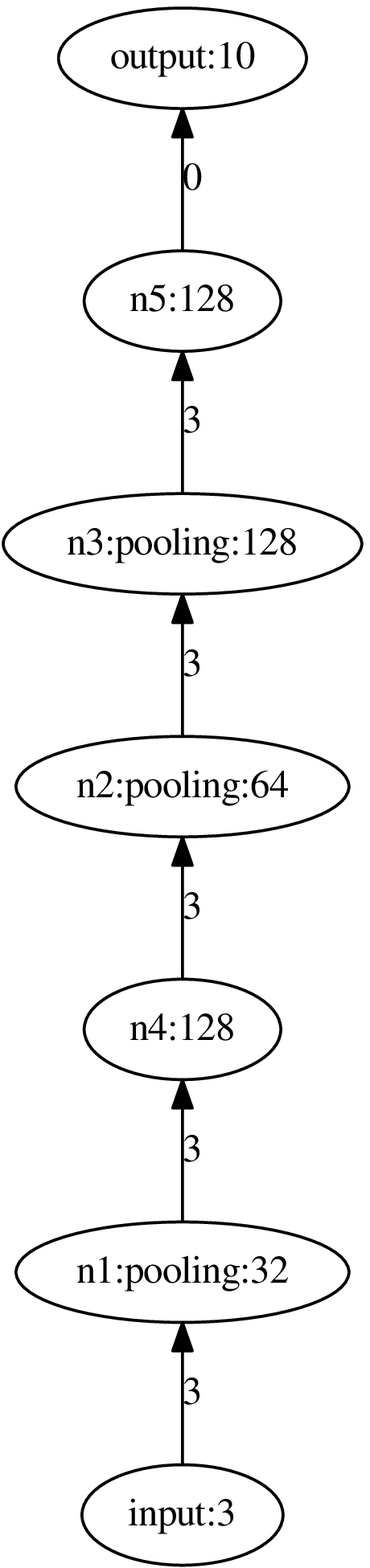} & \includegraphics[scale=0.25]{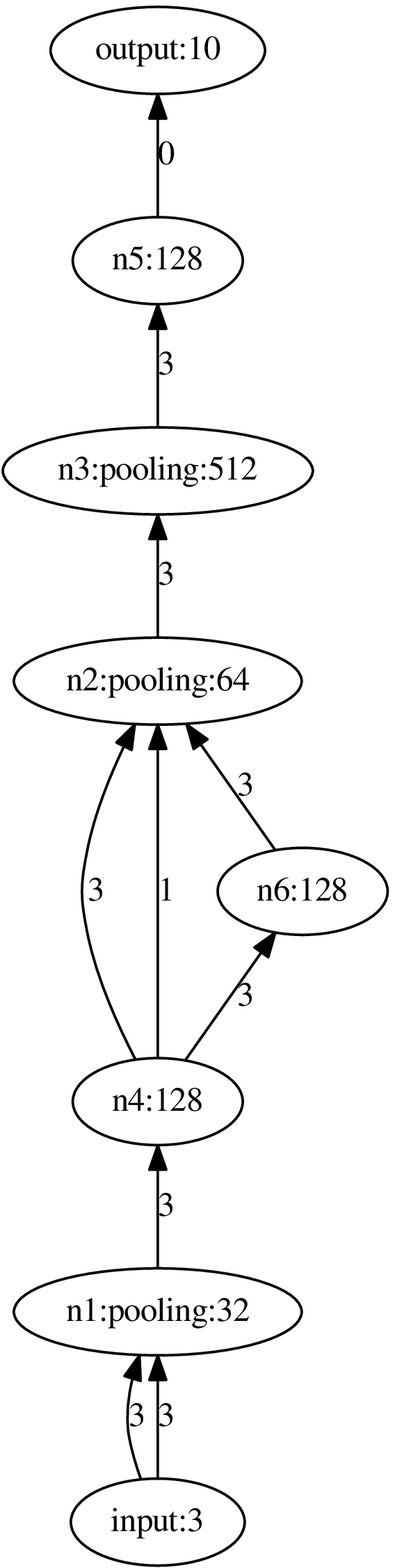} & \includegraphics[scale=0.25]{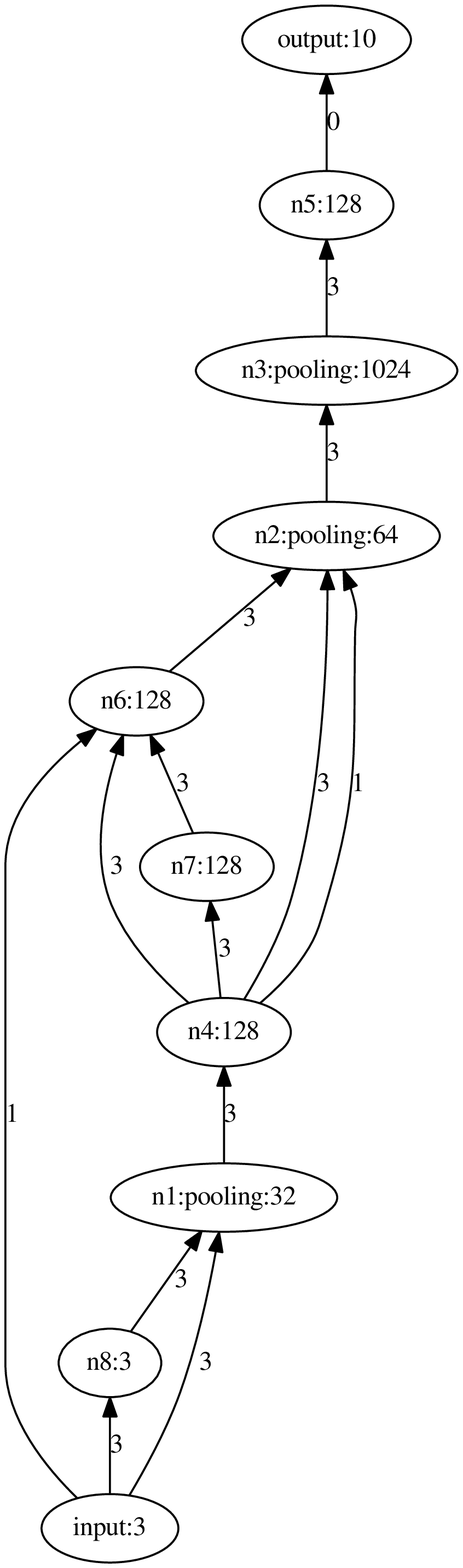}\tabularnewline
(18) & (25) & (29)\tabularnewline
\end{tabular}
\par\end{centering}
\caption{Evolution history of the topology (a) in Fig. \ref{fig:top_networks}.
The numbers below the networks are the numbers of generations. \label{fig:evolve_history}}
\end{figure}

Fig. \ref{fig:evolve_history} shows the topologies of some ancestors
of the best-performing individual during the evolution (individual
$a$ in Fig. \ref{fig:top_networks}).

It is obvious that the network topology simply evolves by increasing
the depth (the longest distance between the source node and the sink
node) in the early stage of the evolution. The network has a layered
structure that there is almost no cross-layer links. This explains
the observations shown in Table \ref{tab:evolved_graph_properties}
that the longest distance and the shortest distance of the evolved
networks are significantly greater than the random evolved networks. 

It is found that the network $18$ in Fig. \ref{fig:evolve_history}
is the common ancestor of the top 1000 best-performing individuals.
This aligns well with the progress of the research in deep CNN that
the accuracy improves by simply using a deeper network structure,
for example: LeNet (5 layers, 1990), AlexNet (8 layers, 2012), VGG
Net (16 and 19 layers, 2014). However, after network $18$ (the 6-layer
network), the depth increasing slows down. The evolution goes towards
adding more links to increase the density and the connectivity of
the networks. 

\subsubsection{Connectivity of a deep CNN}

According to Table \ref{tab:evolved_graph_properties}, the average
generation of the top 1000 individuals is 33.2, much larger than the
longest distance of the network which has the value of 8.67. Note
that the 6-layer network 18 (the 18th generation) in Fig. \ref{fig:evolve_history}
is the common ancestor of the top 1000 individuals. This indicates
that the depth of the network in the last 15 generations only increased
by 2.67, comparing to 4 in the first 18 generations. 

The density and the algebraic connectivity of network 18 is 0.142
and 0.198 respectively. Obviously, when a network reaches to a certain
depth, the density and connectivity becomes a critical factor that
affect the performance. We also noticed that the average of the shortest
distance between the source node and the sink node is 5.60, which
is close to the depth of network 18. This shows that the shortest
distance between the source node and the sink node stays at 6 even
the network gets complicated. This interestingly coincides the phenomenon
of ``six degrees of separation'' that is well-known in social networks
and other types of networks \cite{newman2010networks}. 

Gradient vanishing is a commonly accepted explanation of the difficulty
in training deep CNNs \cite{hochreiter2001gradient}. In backpropagation,
the weights of the deep CNN is updated using gradient descend and
the gradients are backpropagated during training. If a network is
deep, the gradients of some weights may shrink and be close to zero
thus prevent the network from updating. Different approaches have
been invented to overcome or ease the problem. One of the major approaches
is the shortcut links between layers, for example, Residual Net \cite{he2015deepresidual},
DenseNet \cite{huang2016densely} and FractalNet \cite{larsson2016fractalnet}.
The links between the layers actually shorten the path from the source
node to the sink node. This effectively avoids the difficulties of
training a deep CNN caused by gradient vanishing. The results show
that this technique is selected naturally by the evolution. 

\subsubsection{Efficiency of a deep CNN}

As stated in Section \ref{sec:Methodology}, the evolution is in favor
of the networks that are light and learn fast. The top-performing
networks shows the following patterns that can reduce the computational
complexity. 
\begin{itemize}
\item The first node that is connected to the source node is always a pooling
node. This lowers the feature map size and greatly reduces the computational
complexity of the network.
\item Heavy nodes (with large number of channels) are always on the top
of the network. The heaviest node is always above at least two pooling
nodes.
\item Kernel size 3 and 1 are dominant. Since the computational complexity
of the convolution is proportional to the square of the kernel size,
small kernel sizes are preferred.
\end{itemize}
These graph pattern are also very commonly used in human designed
networks \cite{krizhevsky2012imagenet,szegedy2014goingdeeper,he2015deepresidual,huang2016densely,iandola2016squeezenet}. 

As stated in Section \ref{subsec:Knowledge-inheritance-and} and \ref{subsec:Computational-Constraints},
the evolution encourages networks that learn fast. To validate this,
we trained the top 10 networks using the training related hyperparameters
learned during evolution, random initialization and 400 epochs. All
the evolved networks get validation accuracy above 80\% within the
first 6 epochs, which is clearly faster than 12 epochs that is reported
in \cite{miikkulainen2017evolving} where a different evolutionary
strategy was used. 

\subsection{Performance evaluation of the evolved topologies\label{subsec:Performance-of-constructed}}

Because of the limitation of the computing resources during the evolution,
the networks were only trained with one epoch. The accuracies shown
in Fig. \ref{fig:evolution_accuracy} does not reflect the true potentials
of the evolved topologies. We took the top 10 evolved networks and
completed the training using random initialization and 400 epochs.
The training related hyperparameters are learned during the evolution
as described in Section \ref{subsec:Knowledge-inheritance-and}. Table
\ref{tab:top_10_evolved_topologies} shows the validation error rate
of these networks. The values reported are the median values of 5
experiments with random initialization. 

\begin{table*}
\caption{Error rate of the top 10 evolved topologies when fully trained \label{tab:top_10_evolved_topologies}}

\def\arraystretch{0.9} 
\setlength{\tabcolsep}{2pt} 
\centering{}%
\begin{tabular}{ccccccccccc}
\hline 
topology & 1 & 2 & 3 & 4 & 5 & 6 & 7 & 8 & 9 & 10\tabularnewline
\hline 
\hline 
depth & 10 & 9 & 8 & 8 & 9 & 9 & 9 & 9 & 8 & 9\tabularnewline
shortest dist. & 6 & 7 & 4 & 6 & 7 & 4 & 7 & 6 & 4 & 6\tabularnewline
error rate & 7.84\% & 7.72\% & 7.50\% & 7.63\% & 7.37\% & 7.65\% & 7.54\% & 8.07\% & 7.41\% & 7.66\%\tabularnewline
\hline 
\end{tabular}
\end{table*}

Table \ref{tab:top_10_evolved_topologies} shows that the top-performing
networks perform similarly with different network topologies. Note
that none of the shortest distance of the these graphs is greater
than 7. 

\begin{figure}[htbp]
\def\arraystretch{0.8} 
\begin{centering}
\textsf{\includegraphics[scale=0.25]{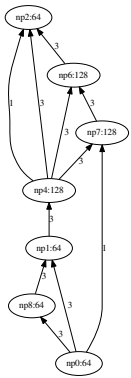}}
\par\end{centering}
\caption{Basic building block for our designed networks \label{fig:building_block}}
\end{figure}

As discussed earlier, we terminated the evolution when the networks
became very complicated. The evolution has not found to the optimal
topologies. However, we can design new topologies using the knowledge
we gained as we discussed in Section \ref{subsec:Statistics-of-top}.
From the topology 1 in Fig. \ref{fig:top_networks}, we take the induced
subgraph that contains the source node, $n1$, $n2$, $n4$, $n6$,
$n7$ and $n8$ as the building block. To keep the subgraph symmetric,
we set the channel of the source node and $n1$ to be 64. Fig \ref{fig:building_block}
shows the building block. We construct the following networks:
\begin{itemize}
\item EVO-44: We stack 7 building blocks by connecting the output of a lower
block to the input of the upper block. Pooling is applied to output
node of the 2nd, the 4th and the 6th block. The depth of this network
is 44, thus named EVO-44. The shortest distance of the graph is 23. 
\item EVO-44a: We stack 7 building blocks similarly as EVO-44. To reduce
the distance between the source node and the sink node, we add edges
to connect $n7$ to the $n8$ in the adjacent upper block, and $n1$
to $n6$ in the adjacent upper block. If the sizes of the feature
map of the two connected nodes are different, a suitable stride is
used. With these extra edges, the shortest distance of the graph 13. 
\item EVO-44b: We stack 7 building blocks in the same way as EVO-44. We
connect $n7$ to the all $n8$ in the upper blocks and $n1$ to all
$n6$ in all the upper blocks. The shortest distance of the graph
is 5. 
\item EVO-91: We stack 15 building blocks in the same way as EVO-44. The
pooling is applied to the output node of the 4th, the 8th and the
12th block. The depth of this deep CNN is 91. The shortest distance
of the graph is 47.
\item EVO-91a: We build the deep CNN similar to how EVO-44a is built except
that 15 build blocks are used. The pooling is applied to the output
of the 4th, the 8th and the 12th block. The shortest of the graph
is 25.
\item EVO-91b: We build the deep CNN similar to how EVO-44b is built except
that 15 build blocks are used. The pooling is applied to the output
of the 4th, the 8th and the 12th block. The shortest distance of the
graph is 5.
\end{itemize}

We evaluated our designed networks on CIFAR-10, CIFAR-100 and SVHN
dataset. On the two CIFAR datasets, the standard image augmentation
(flipping and clipping) is used. The maximum number of epoch is 400.
Training related hyperparameters are learned during evolution as specified
in Section \ref{subsec:Evoluation}: Adam \cite{kingma2014adama}
optimization method is used; initial learning rate is 0.0005; no dropout
is used; no weight decay is used. Learning rate is reduced by a factor
of 5 on every 100 epochs. On SVHN dataset, no data augmentation is
applied. The maximum number of epoch is 180. Learning related hyperparameters
are same as those used for CIFAR datasets. Learning rate is reduced
by a factor of 5 on every 60 epochs. 

Table \ref{tab:error_rate_of_constructed} shows the error rate of
our designed networks tested on CIFAR-10, CIFAR-100 and SVHN datasets.
All values reported on our topologies are the median values 5 experiments
with random initialization. Error rate of some other human designed
network are also shown for comparison. 

\begin{table}
\caption{Comparison of the error rate of our networks and other networks on
CIFAR-10, CIFAR-100 and SVHN datasets. \label{tab:error_rate_of_constructed}}

\def\arraystretch{0.9} 
\setlength{\tabcolsep}{2pt} 
\centering{}%
\begin{tabular}{cccc}
\hline 
 & CIFAR-10 & CIFAR-100 & SVHN\tabularnewline
\hline 
\hline 
NIN & 8.8\%\cite{zagoruyko2016wideresidual} & 35.7\%\cite{zagoruyko2016wideresidual} & 2.35\cite{huang2016densely2}\tabularnewline
DSN & 8.2\%\cite{zagoruyko2016wideresidual} & 34.6\%\cite{zagoruyko2016wideresidual} & 1.92\cite{huang2016densely2}\tabularnewline
Highway & 7.5\%\cite{zagoruyko2016wideresidual} & 32.4\%\cite{zagoruyko2016wideresidual} & -\tabularnewline
All-CNN\cite{springenberg2014striving} & 7.25\% & 33.71\% & -\tabularnewline
ResNet-56\cite{he2015deepresidual} & 6.97\% & - & -\tabularnewline
ResNet-110\cite{he2015deepresidual} & 6.43\% & 25.2\%\cite{zagoruyko2016wideresidual} & 2.01\cite{huang2016densely2}\tabularnewline
pre-act-ResNet-110\cite{he2016identity} & 6.37\% & - & -\tabularnewline
DenseNet-100\cite{huang2016densely2} & 3.74\% & 19.3\% & 1.59\%\tabularnewline
\hline 
EVO-8 & 7.84\% & 30.0\% & 2.47\%\tabularnewline
EVO-44 & 5.99\% & 29.0\% & 2.03\%\tabularnewline
EVO-44a & 5.74\% & 26.8\% & 1.99\%\tabularnewline
EVO-44b & 5.65\% & 25.8\% & 1.97\%\tabularnewline
EVO-91 & 6.35\% & 34.0\% & 2.02\%\tabularnewline
EVO-91a & 5.71\% & 28.7\% & 1.96\%\tabularnewline
EVO-91b & 5.19\% & 24.6\% & 1.85\%\tabularnewline
\hline 
\end{tabular}
\end{table}

The results in Table \ref{tab:error_rate_of_constructed} show that
extra links that connect the nodes between different blocks improve
the performance. Due to the large shortest distance between the source
node and the sink node, the accuracy of EVO-91 is clearly worse than
the other networks. The EVO-91b achieves the best performance by increasing
the depth of the network and keeps the shortest distance of the source
node and sink node low. 

Comparing to Residual Net model, the constructed topologies achieve
better results with less layers. It should be noted that DenseNet
gives better performance with a much higher graph density. However,
it should be noted that the shortest distance between the source node
and the sink node of the DenseNet-100 is 6, which is also coordinate
to the top-performing network during the evolution as discussed in
Section \ref{subsec:Statistics-of-top}. 

\subsection{Limitations}

First, even though the knowledge inheritance greatly alleviates the
difficulties of training a deep CNN, computing resource is significantly
limit the scope of our evolution. We had to stop the training of each
individual at very early stage (one epoch). This becomes more severe
when networks are getting complicated. 

Second, our evolution is based on the canonical neuron model. This
is sufficient to study the network topology, but not likely to find
the best-performing deep CNN. For example, previous research showed
that pre-activation structure, which separates the propagation and
activation in Eq. \ref{eq:neuron_model}, can greatly improve the
performance of Residual Net \cite{he2016identity}. Zoph et al. evolved
a network with many different neuron models and achieved the state-of-the-art
results on CIFAR-10 and ImageNet dataset \cite{zoph2017learning}.

Our evolutionary algorithm evolves slowly when network topology is
getting big and complicated. For example, the changes generated from
the mutations defined in Section \ref{subsec:Mutations} is relatively
minor when the network is big. More advanced mutations and other techniques
are required to increase the exploration when the evolution reaches
to a certain stage. 

\section{Conclusion\label{sec:Conclusion}}

In this paper, we present an evolutionary algorithm to find better
deep CNN topologies. Our evolutionary strategy is based on asexual
reproduction. One of the 5 predefined mutations is applied to a parent
to generate an offspring. Boltzmann distribution is used for the stochastic
rank-proportional selection strategy. To deal with the difficulty
of training deep CNNs, we exploit the concept of knowledge inheritance
and each individual is trained with one epoch. Training related hyperparameters
are learned during the evolution in virtue of the concept of knowledge
learning. 

We applied the proposed algorithm to the image classification problem
using CIFAR-10 dataset and studied the topologies that performed well
during the evolution. Our studies verify some generally accepted techniques
in human designed networks. The studies also show that the shortest
distance from the source node to the sink node, graph density and
graph connectivity are also major factors that affect the performance
of a deep CNN. Based on these findings, we designed new topologies
and evaluated their performance using CIFAR-10, CIFAR-100 and SVHN
datasets. The experiment results confirm the efficiency of the guidelines
we learned from the evolution. Our designed topologies can outperform
Residual Net with less layers. 

We also discussed the limitations of our evolution experiments. Although
our evolutionary algorithms employ different techniques to speed up
the training of deep CNNs, computing resource still greatly hinder
the evolution since complex topologies cannot be effectively evaluated.
Also better evolutionary strategy should be considered when the topologies
are getting complicated, similar to nature where complex organisms
always adapt complicated reproduction mechanism. This will be the
main direction of our future research. We make the software package
and all the topologies that had been evaluated during the evolution
publicly available.\footnote{ <URL to be released later>.} We hope
our research will be beneficial for the society to get better understandings
of the topologies for the deep CNNs and design more efficient networks.

\bibliographystyle{plain}
\bibliography{evo}

\end{document}